\documentclass[conference]{IEEEtran}
\IEEEoverridecommandlockouts
\usepackage[T1]{fontenc}    
\usepackage{cite}
\usepackage{graphicx}
\usepackage{amsmath}
\usepackage{amssymb}
\usepackage{booktabs}
\usepackage{subcaption}

\usepackage{tabularx}
\usepackage{adjustbox}
	
\usepackage{color, colortbl}
\usepackage[normalem]{ulem}

\usepackage[natural]{xcolor}
\usepackage[markup=underlined]{changes}
\usepackage{todonotes}
\usepackage{mathtools}
\usepackage{url}
\usepackage{algorithm}
\usepackage{algorithmicx}
\usepackage[noend]{algpseudocode}
\usepackage{gensymb}
\usepackage{siunitx} 
\usepackage{interval}
\usepackage[pagebackref,breaklinks,colorlinks]{hyperref}
\usepackage{svg}
\usepackage{mathrsfs}
\usepackage{amssymb}
\usepackage{pifont}
\usepackage{stfloats} 
\newcommand{\xmark}{\ding{55}}%
\usepackage{pdflscape}
\usepackage{afterpage}
\usepackage{multirow}
\newcommand{\STAB}[1]{\begin{tabular}{@{}c@{}}#1\end{tabular}}

\newcommand\copyrighttext{%
  \footnotesize \textcopyright 2025 IEEE. Personal use of this material is permitted.
  Permission from IEEE must be obtained for all other uses, in any current or future
  media, including reprinting/republishing this material for advertising or promotional
  purposes, creating new collective works, for resale or redistribution to servers or
  lists, or reuse of any copyrighted component of this work in other works.}

\newcommand\copyrightnotice{%
\begin{tikzpicture}[remember picture,overlay]
\node[anchor=south,yshift=10pt] at (current page.south) {\fbox{\parbox{\dimexpr\textwidth-\fboxsep-\fboxrule\relax}{\copyrighttext}}};
\end{tikzpicture}%
}

\newcommand{\norm}[1]{\left\lVert#1\right\rVert}

\title{Real Time Semantic Segmentation of High Resolution Automotive LiDAR Scans}

\author{Hannes Reichert, Benjamin Serfling, Elijah Schüssler, Kerim Turacan, Konrad Doll, and Bernhard Sick
	\thanks{H. Reichert, B. Serfling, E. Schüssler, K. Turacan. and K. Doll are with the Faculty of Engineering,
		University of Applied Sciences Aschaffenburg, Aschaffenburg, Germany
		{\tt\footnotesize firstname.lastname@th-ab.de}}
	\thanks{B. Sick is with the Intelligent Embedded Systems Lab, University of Kassel,
		Kassel, Germany
		{\tt\footnotesize bsick@uni-kassel.de}}
}

\begin{document}

\maketitle
\copyrightnotice
\begin{abstract}
In recent studies, numerous previous works emphasize the importance of semantic segmentation of LiDAR data as a critical component to the development of driver-assistance systems and autonomous vehicles. However, many state-of-the-art methods are tested on outdated, lower-resolution LiDAR sensors and struggle with real-time constraints. This study introduces a novel semantic segmentation framework tailored for modern high-resolution LiDAR sensors that addresses both accuracy and real-time processing demands. We propose a novel LiDAR dataset collected by a cutting-edge automotive 128 layer LiDAR in urban traffic scenes.
Furthermore, we propose a semantic segmentation method utilizing surface normals as strong input features. Our approach is bridging the gap between cutting-edge research and practical automotive applications. Additionaly, we provide a Robot Operating System (ROS2) implementation that we operate on our research vehicle. Our dataset and code are publicly available: \href{https://github.com/kav-institute/SemanticLiDAR}{https://github.com/kav-institute/SemanticLiDAR}.

\begin{figure}[!h] 
\vspace{-2mm} 
    \centering
    \includegraphics[width=\columnwidth]{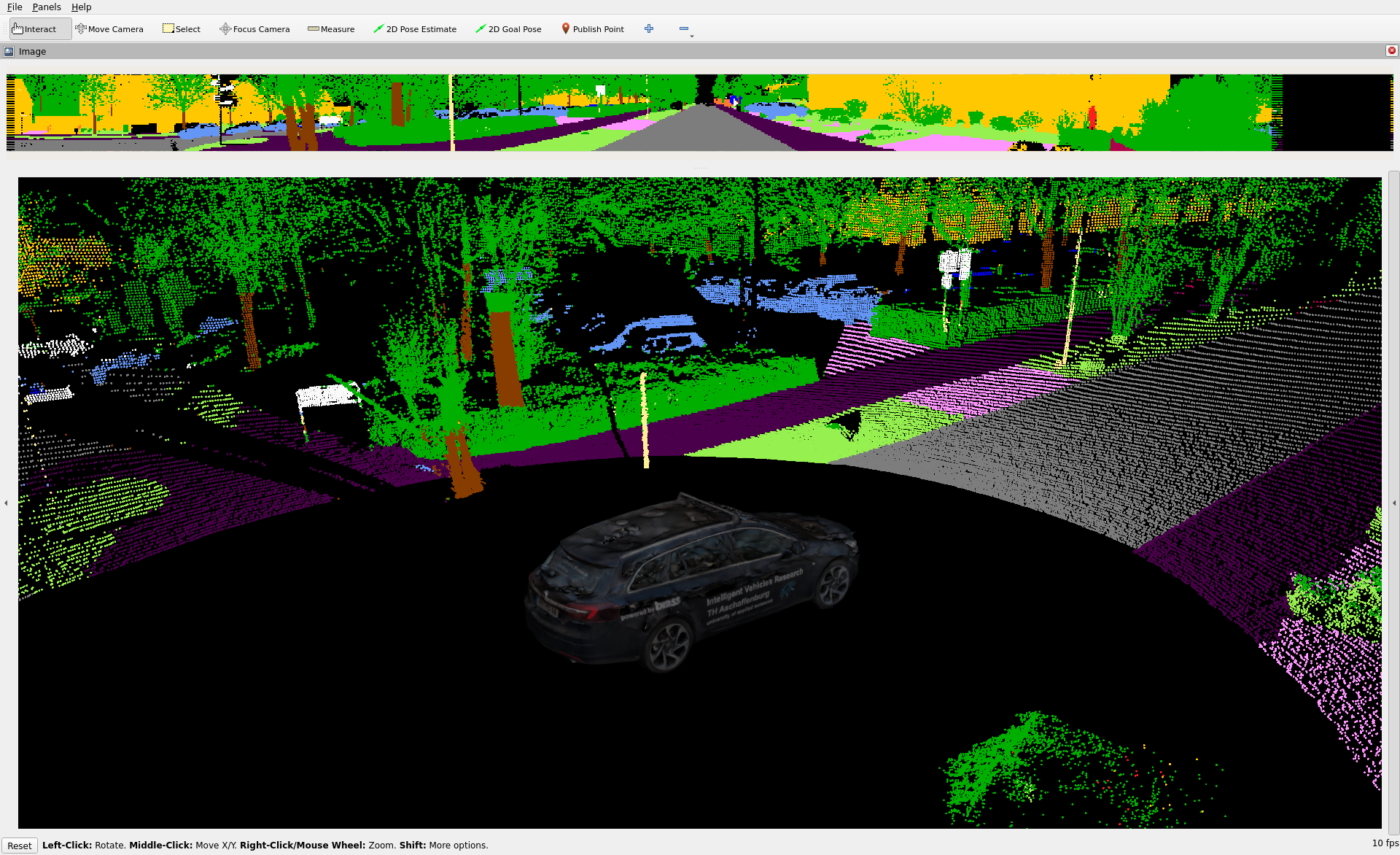}
    \caption*{\textit{\textbf{Visual Abstract}}: Semantic segmentation of a LiDAR scan in RVIZ (ROS2).}
    \label{fig:visual_abstract}
    \vspace{-5mm} 
\end{figure}

\end{abstract}
\section{\large Introduction}
\label{sec_introduction}
\noindent LiDAR sensors play a vital role in autonomous vehicle perception systems, providing precise and reliable geometric data through 3D point clouds. Semantic segmentation assigns class labels to each point in the cloud, enabling scene understanding. Recent advances in LiDAR technology have significantly improved measurement accuracy, resolution, and calibration.
However, many existing segmentation methods have not kept up with advancements in hardware, often relying on outdated, low-resolution sensors. This underuse of modern LiDAR sensors poses a critical challenge as it hinders technological advancement. Additionally, evolving hardware introduces new processing requirements that current methods cannot address, limiting their applicability to autonomous driving products.

\subsection{Related Work} \label{sec_sota}
\noindent\textbf{Datasets:} Several publicly available datasets exist for the semantic segmentation of LiDAR scans. Starting with SemanticKITTI \cite{behley2019iccv} from 2019, followed by SemanticPOSS \cite{pan2020semanticposs} and SemanticUSL \cite{jiang2020lidarnet} in 2020 and nuScenes \cite{fong2021panoptic} in 2021. Waymo Open Dataset \cite{Waymo} was released in 2020 and updated with LiDAR semantic segmentation labels in 2022. SemanticSTF \cite{xiao20233d} is the newest dataset released in 2023. SemanticKITTI offers a large-scale dataset with an extensive number of annotated frames but is limited in sequence variability, focusing primarily on suburban driving scenarios. The Waymo Open Dataset and nuScenes offer large-scale datasets. SemanticPOSS, SemanticUSL, and SemanticSTF are relatively small datasets designed for specific use cases, such as domain transfer or robustness evaluation under adverse weather conditions. Only SemanticKITTI, SemanticUSL, and SemanticSTF utilize the same label definitions. In \autoref{tab:dataset} we list some statistics for these datasets and the sensors used. Dataset that share SemanticKITTI label definitions are underlined in the \#Classes column. Note that the sensors used in all datasets are low-resolution compared to modern standards.
However, dataset that use modern high-resolution sensors are essential to align with the developments in sensor hardware.
\begin{table}[h]
    \centering
    \caption{Dataset Comparison}
    \label{tab:dataset}
\resizebox{\columnwidth}{!}{
\begin{tabular}{|l|l|l|l|l|l|}
\toprule
              & Sensor                                                       & \#Layer & \#Scans & FPS   & \#Classes                                         \\
\midrule
SemanticKITTI \cite{behley2019iccv} & Velodyne HDL64 & 64      & 43,552  & 10    & \underline{28} \\
SemanticPOSS \cite{pan2020semanticposs} & Hensai Pandora                                               & 40      & 2,988   & 10    & 14                                                \\
SemanticUSL \cite{jiang2020lidarnet}  & Ouster OS1 (Rev5)                                             & 64      & 1,200   & 10    & \underline{20} \\
SemanticSTF \cite{xiao20233d}  & Velodyne HDL64 & 64      & 2,076   & \xmark & \underline{20}                        \\
Waymo  \cite{Waymo}       & in-house LiDAR                                               & 64      & 46,000  & 2     & 23                                                \\
nuScenes   \cite{fong2021panoptic}   & Velodyne HDL32E                                              & 32      & 40,000  & 2.5   & 32                                                \\
\textbf{SemanticTHAB (our)}  & Ouster OS2 (Rev7)                                            & 128     & 4,750   & 10    & \underline{20}                        \\
\bottomrule
\end{tabular}}
\end{table}

\noindent\textbf{Real-Time Segmentation of LiDAR Point Clouds:} Since the release of the SemanticKITTI benchmark in 2019 \cite{behley2019iccv}, numerous semantic segmentation methods have emerged for LiDAR Point Clouds. These approaches utilize various data representations, balancing accuracy and computational efficiency. For example, FRNet \cite{FRNet}, FIDNet \cite{Zhao2021FIDNetLP} and CENet \cite{cheng2022cenet} employ spherical projection and image-based processing, such as Convolutional Neural Networks (CNNs), to achieve high frames per second (FPS) and strong mean Intersection over Union (mIoU) scores. Vision transformer (ViT)-based models like RangeFormer \cite{RangeFormer} and RangeViT \cite{RangeViT} can surpass these mIoU scores, but typically operate at lower frame rates. Other methods, including SphereFormer \cite{SphereFormer}, Cylinder3D \cite{Cylinder3D}, and MinkUNet \cite{tang2020searching}, utilize 3D representations such as voxels or raw point clouds. Although these approaches achieve high mIoU scores, they often lack real-time capability due to lower frame rates. Currently, the model with the highest performance in terms of mIoU is the Point Transformer V3 foundation model \cite{wu2024ptv3}. \autoref{tab:benchmark_sup} compares various methods, listing their FPS and mIoU scores on the SemanticKITTI validation set.

\begin{table}[h]
\caption{Benchmark Comparison}
\centering
\scalebox{0.75}{
\begin{tabular}{|r||r|r|c|c|}
    \toprule
    & \textcolor{darkgray}{Representation} & \textcolor{darkgray}{Parameters}  & \textcolor{darkgray}{\textbf{FPS}~{\small$\uparrow$}} & \textcolor{darkgray}{\textbf{mIoU}~{\small$\uparrow$}} 
    \\
    \midrule
    MinkUNet~\cite{tang2020searching} & Sparse Voxel & $21.7$ M & $9.1$ & $62.8$ 
    \\
    Cylinder3D~\cite{Cylinder3D} & Sparse Voxel & $55.9$ M & $6.2$ & $65.9$ 
    \\
    WaffleIron~\cite{waffleiron} & Raw Points & $6.8$ M & $7.4$ & $68.0$
    \\
    PTV3~\cite{wu2024ptv3} & Raw Points & \xmark & \xmark & $72.3$
    \\
    RPVNet~\cite{RPVNet}  & Multi-View& $24.8$ M & $7.9$ & $65.5$ 
    \\
    FIDNet~\cite{Zhao2021FIDNetLP}  & Range View & $6.1$ M & $31.8$ & $58.9$ 
    \\
    CENet~\cite{cheng2022cenet}+~\cite{pvkd} & Range View & $6.8$ M & $33.4$ & $67.6$ 
    \\
    RangeFormer~\cite{RangeFormer}  & Range View  & $24.3$ M & $6.2$ & $67.6$ 
    \\
    FRNet~\cite{FRNet}  & Range View  & $7.5$ M & $33.8$ & $67.1$ 
    \\
    RangeVit~\cite{RangeViT}  & Range View  & $23.7$ M & 10.0 & $60.7$ 
    \\
    \textbf{Our @} $64 \times 512$  & Range View  & $25.1$ M & $85.21$ & $63.71$ 
    \\\bottomrule
\end{tabular}}
\label{tab:benchmark_sup}
\end{table}
\noindent The FPS values are sourced from \cite{FRNet}, which reports the results using a single GeForce RTX 2080Ti GPU.  Note that the range view methods benchmarked in \cite{FRNet} use a resolution of $64 \times 512$. Thus, we assume that the FPS measures from \cite{FRNet} are lower when applied to the high-resolution LiDAR used in our study. For a deployed system, we assume a tolerable inference time of 50 ms (20+ FPS). However, since segmentation is an early block in a full perception pipeline, we halve the inference time to 25 ms (40+ FPS) to free up capacity for decision systems.
We conclude that CNN architectures and range view representation give a suitable trade-off between accuracy and real-time performance. We see CENet as a strong baseline with a high mIoU ($67.6$) while being close to our inference time requirement with $33.4$ FPS.

\subsection{Main Contributions} \label{sec_contrib}
\noindent The related work section presents various approaches and datasets focused on the semantic segmentation of LiDAR data. However, it is evident that both methods and datasets have limitations regarding the sensors' resolution, their relevance to current hardware advancements, and the runtime efficiency of many existing approaches. In this context,
the main contributions of this paper can be summarized as
follows:
\begin{itemize}
    \item A dataset for semantic segmentation of high-resolution automotive LiDAR data.
    \item A real-time capable method for semantic segmentation utilizing surface normals as additional input features.
    \item Publicly available code for training, testing, and a ROS2 demonstration system.
\end{itemize}
\section{\large Dataset}
\label{sec:dataset}
\begin{figure}[!h] 
    \centering
    \includegraphics[width=\columnwidth]{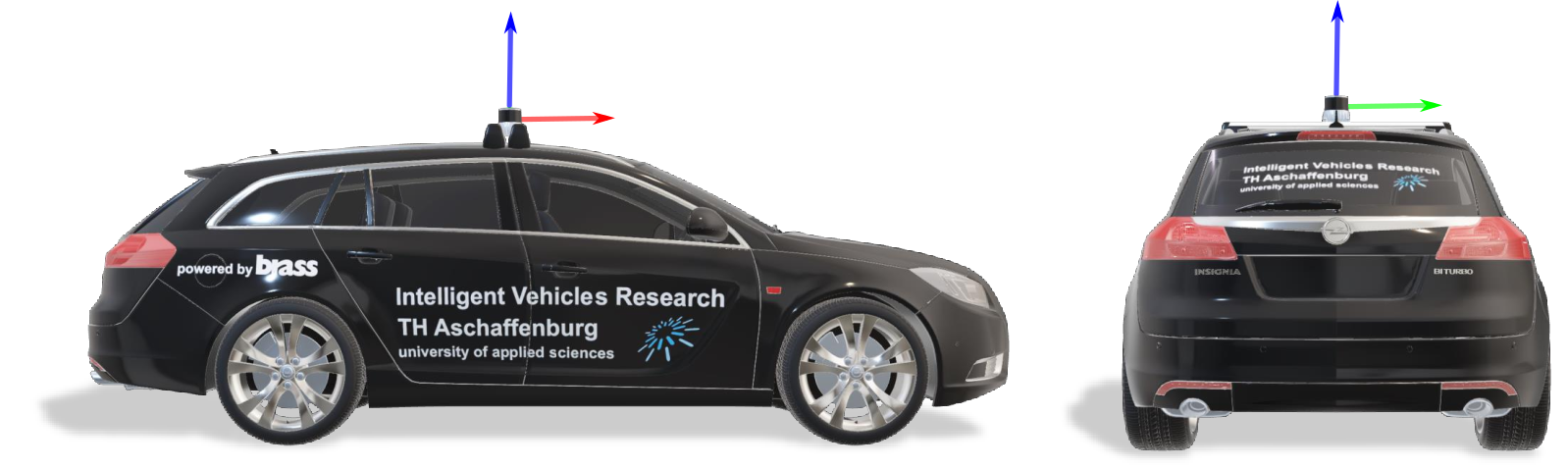}
    \caption{\textit{\textbf{System Setup}}: We use an Ouster OS2-128 (Rev7) LiDAR, mounted on the rooftop of a research vehicle.}
    \label{fig:sensor_setup}
    \vspace{-3mm} 
\end{figure}

\begin{figure}[!h] 
    \centering
    \includegraphics[width=\columnwidth]{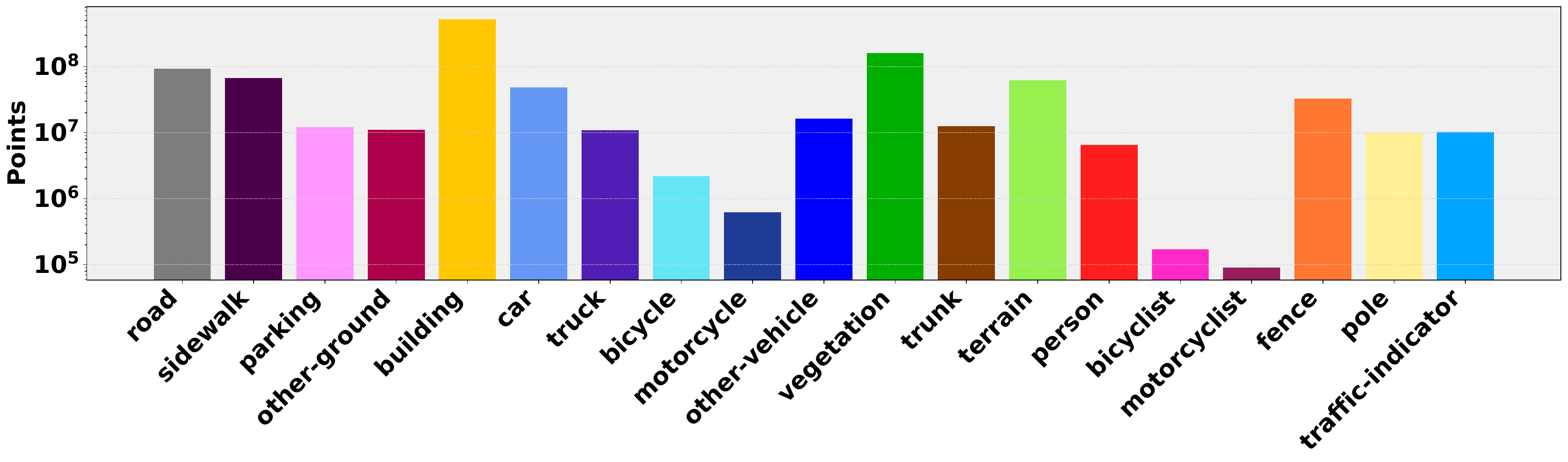}
    \caption{Distributions of semantic classes in SemanticTHAB.}
    \label{fig:SemanticTHAB_statistics}
\vspace{-3mm} 
\end{figure}

\begin{figure*}[htb] 
    \centering
    \includegraphics[width=\textwidth]{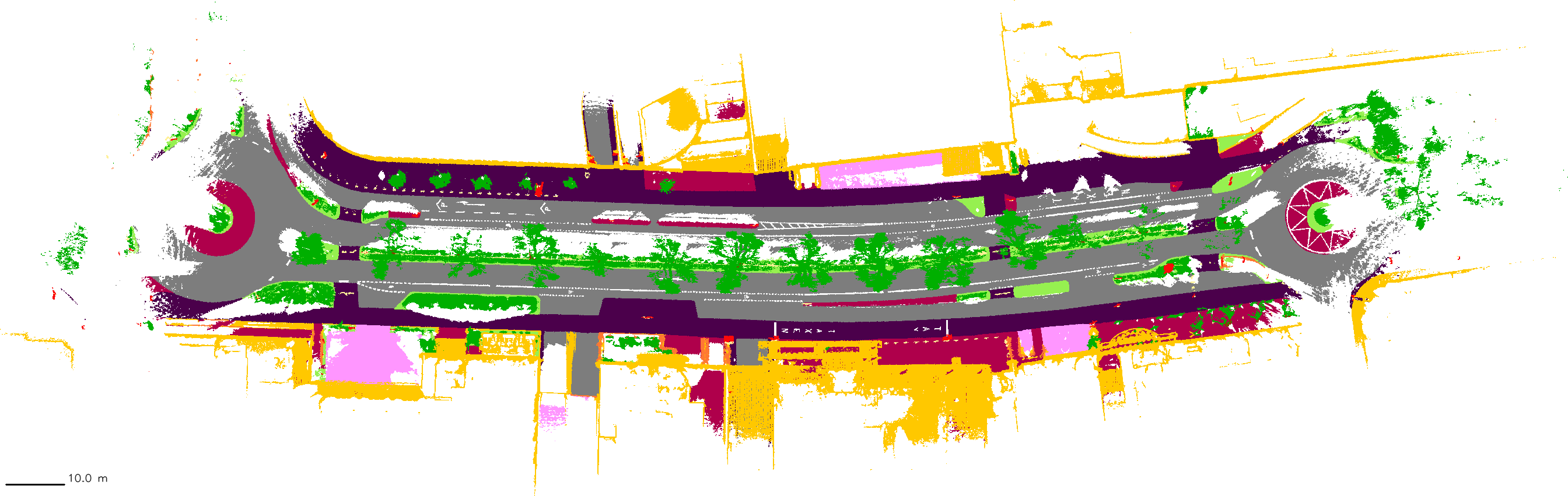}
    \caption{Reference map of scene \textbf{0002}. Dynamic objects are removed for better visibility.}
    \label{fig:map}
    \vspace{-5mm} 
\end{figure*}

\noindent Our sensor setup is illustrated in \autoref{fig:sensor_setup}. It features a modern, automotive-grade LiDAR mounted on the rooftop of a research vehicle, equipped with hardware capabilities for data recording and processing.
For the annotation process, we utilize \textit{PointLabeler} by J. Behley et al. \cite{behley2019iccv}, which supports the grouping and annotation of multiple LiDAR scans through ego motion compensation. To estimate ego-motion, we employ the SLAM module from the Ouster SDK, which incorporates a high-level implementation of \textit{KISS-ICP} \cite{kiss-icp}. We adopt the class definitions from SemanticKITTI in our annotation process. Note that we renamed the class \textit{traffic-sign} to \textit{traffic-indicator}, grouping \textit{traffic-signs} and \textit{lane-markings} under this unified category. In SemanticKITTI, \textit{lane-markings} are treated as \textit{road} due to their low visibility in the reflectivity measures of the LiDAR sensor used in the dataset. The class distribution in our data set is illustrated in \autoref{fig:SemanticTHAB_statistics}, which uses a logarithmic scale to represent the class frequencies. As expected, certain classes such as buildings, vegetation, and roads dominate the dataset, reflecting the environmental conditions of the recorded locations. Additionally, the person class appears relatively frequently, aligning with our focus on pedestrian-rich areas. 
We recorded scenes in diverse urban environments, emphasizing variation in both surroundings and road topology, as detailed in \autoref{tab:scenes}. Scene 0002, shown in \autoref{fig:map}, represents a complex urban setting in Aschaffenburg, Germany. 
\begin{table}[!h]
\centering
\caption{Dataset}
    \label{tab:scenes}
\begin{tabular}{|l|l|l|}
\hline
Sequence & Scans & Location                  \\ \hline
0000     & 1090  & Residential \& Industrial \\ \hline
0001     & 344   & City Ring Road            \\ \hline
0002     & 228   & Inner City                \\ \hline
0003     & 743   & Pedestrian Area           \\ \hline
0004     & 400   & Inner City                \\ \hline
0005     & 603   & Inner City                \\ \hline
0006     & 320   & Inner City                \\ \hline
0007     & 517   & Residential               \\ \hline
0008     & 505   & Campus                    \\ \hline
\end{tabular}
\end{table}

\section{\large Method}\label{sec_method}
\begin{figure}[!h]
    \centering
    \begin{minipage}{\columnwidth}
        \centering
        \includegraphics[width=\textwidth]{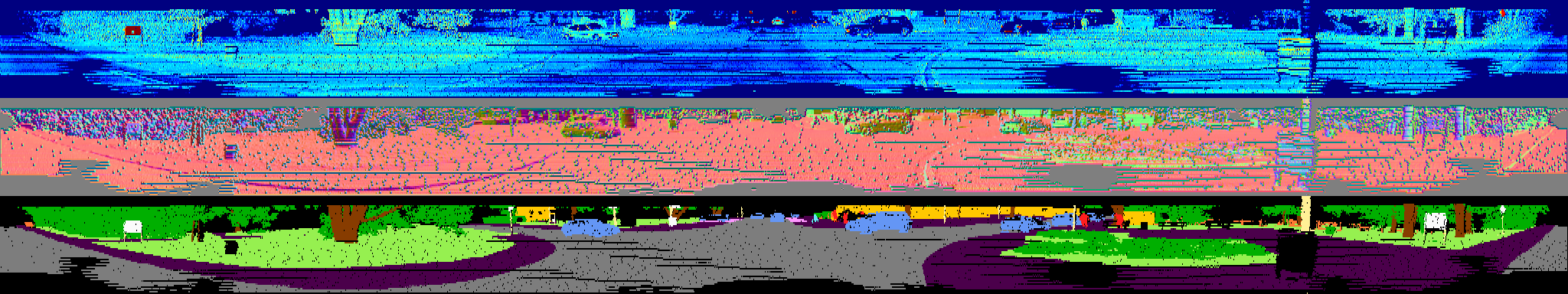}
        \subcaption{SemanticKitti}
        \label{fig:kitti}
    \end{minipage}
    \begin{minipage}{\columnwidth}
        \centering
        \includegraphics[width=\textwidth]{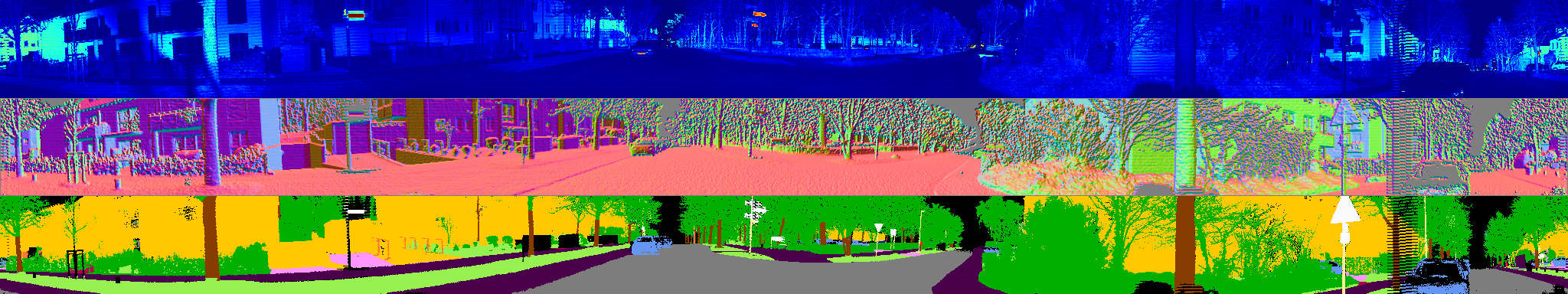}
        \subcaption{SemanticTHAB (Ours)}
        \label{fig:thab}
    \end{minipage}
    \caption{Reflectivity (top), surface normals (middle), and labels (bottom) for SemanticKITTI (a) and SemanticTHAB (b).}
    \label{fig:datasets}
    \vspace{-4mm} 
\end{figure}

\noindent Using a modern sensor such as the Ouster OS2-128 (Rev7) enables us to work with less noisy, better calibrated, and higher resolution data, as well as better reflectivity measures compared to the Velodyne HDL-64E used in SemanticKITTI as shown in \autoref{fig:datasets}. Thanks to its good calibration, no post-processing such as re-projection with K-Nearest-Neighbors (KNN) interpolation is needed, further improving the runtime efficiency of our approach.

\subsection{Spherical Projection}
\label{sec:spherical projection}

\noindent Our approach relies on a staggered image-like representation, where neighboring pixels correspond to adjacent LiDAR measurement rays and their respective measured points, as introduced in \cite{2023.reichert}. This representation is obtained through a spherical image projection.
Each point in the point cloud, defined by its Cartesian coordinates $[x,y,z]^T$, is transformed into spherical coordinates, represented as $[\phi, \theta, r]^T$. The azimuth $\phi$ corresponds to the angle of the point in the $xy$-plane, the inclination $\theta$ denotes the angle from the positive $z$-axis, and $r$ is the distance from the origin. This spherical projection captures the geometry of the sensor in a single image. We then use the following projection model to obtain a pixel $\vec{u}$ of a 3D point in a staggered spherical image representation:
\begin{equation}
\underbrace{\begin{bmatrix}
 u\\
 v\\
 1\\
\end{bmatrix} }_{\vec{u}}
=
\underbrace{\begin{bmatrix}
\frac{1}{\triangle \phi} &  0 &  c_{\phi}\\
 0 &  \frac{1}{\triangle \theta} &  c_{\theta}\\
 0 & 0 & 1\\
\end{bmatrix}}_{\mathbb{K}} \cdot \underbrace{\begin{bmatrix}
 \phi\\
 \theta\\
 1\\
\end{bmatrix}}_{\vec{x}}  
\end{equation}

\noindent Similarly to the projection model of the pinhole camera, the projection matrix $\mathbb{K}$ describes a discretization $\triangle \phi$, $\triangle \theta$ along the angles $\phi$, $\theta$ and a shift of the center coordinates $c_{\phi}$, $c_{\theta}$ defined by the height and width of the resulting image.
Building a spherical image $I_{x,y,z}(u,v)$ with the projection $\pi$ can be described as:
\begin{equation}
I_{x,y,z}(u,v)=\pi(\vec{x}, \mathbb{K})
\end{equation}
Similarly, we can construct $I_{\textit{ref}}(u,v)$ for the reflectivity channel and compute the Euclidean norm to derive $I_{r}(u,v)$ as a range measure from $I_{x,y,z}(u,v)$. 

\subsection{Surface Normal Estimation}
\label{sec:surface normal estimation}
\noindent Surface normals are valuable features in 3D point clouds as they provide information about the orientation and homogenity of objects. This is particularly useful in semantic segmentation for identifying man-made structures such as buildings, detecting curbs, and distinguishing objects that stand out from the ground. 
We use the method described in \cite{surface_normals} to efficiently estimate surface normals from a spherical image $I_{x,y,z}$. The neighborhood of a pixel $P_c=I_{x,y,z}(u,v)$, namely $P_b=I_{x,y,z}(u+1,v)$ and $P_a=I_{x,y,z}(u,v+1)$ in the 3D point cloud can be used with a directional derivative filter over the individual channels of $I_{x,y,z}$ to build the vector $\overrightarrow{P_cP_b}$ for every pixel position $(u,v)$:
\begin{equation}
\underbrace{\begin{bmatrix}
 I_x(u+1,v) - I_x(u,v)\\
 I_y(u+1,v) - I_y(u,v)\\
 I_z(u+1,v) - I_z(u,v)\\
\end{bmatrix}}_{
I_{\overrightarrow{P_cP_b}}(u,v)} = \begin{bmatrix}
(S_u * I_x)(u,v) \\
(S_u * I_y)(u,v) \\
(S_u * I_z)(u,v) \\
\end{bmatrix}
\label{eq:grad_img_u}
\end{equation}
With $S_u$ representing a horizontal gradient filter and $*$ denoting the convolution operator. With $S_v$ as the vertical derivative filter we can build $I_{\overrightarrow{P_cP_a}}(u,v)$ analogous.
 To obtain the surface normals $I_{n_x,n_y,n_z}(u,v)$ we build the cross product over those vectors:
 
 \begin{equation}
    I_{n_x,n_y,n_z}(u,v) = \frac{I_{\overrightarrow{P_cP_b}}(u,v) \times I_{\overrightarrow{P_cP_a}}(u,v)}{\norm{I_{\overrightarrow{P_cP_b}}(u,v) \times I_{\overrightarrow{P_cP_a}}(u,v)}_2}
    \label{eq:cross_img}
\end{equation}


\subsection{Model Architecture}
\label{sec:model architecture}
\begin{figure}[!h] 
    \centering
    \includegraphics[width=\columnwidth]{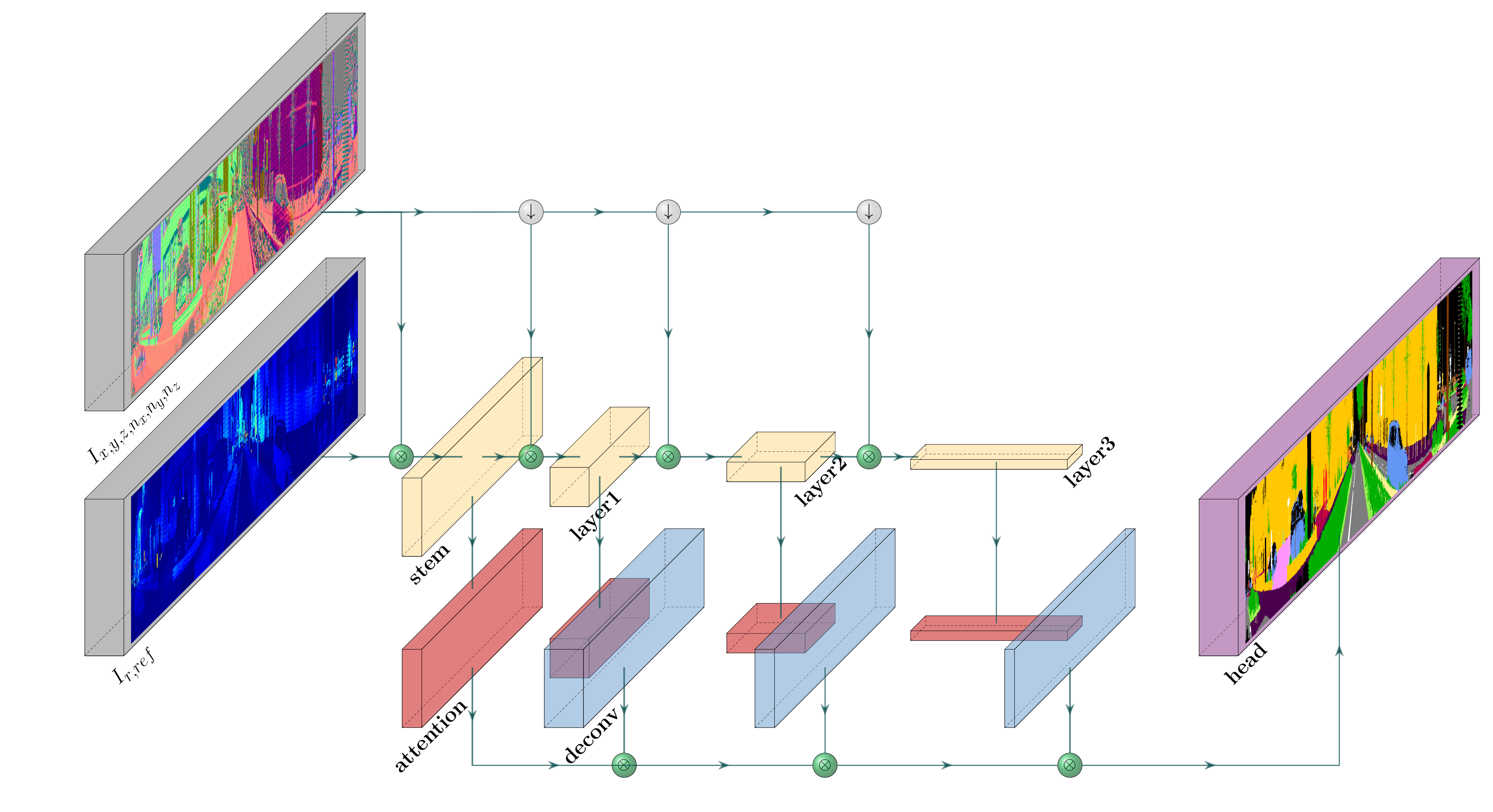}
    \caption{\textit{\textbf{Model Architecture}}: yellow: backbone path, red: attention mechanism, light blue: deconvolution, $\downarrow$: nearest-neighbor downsampling, $\otimes$: concatenation, violett: segmentation head.}
    \label{fig:network_arch}
    \vspace{-5mm} 
\end{figure}
\noindent For our model architecture, we rely on CNNs, leveraging predefined architectures as backbones, with modifications to enable efficient processing of 3D data through specialized convolutional operations. Our model architecture is shown in \autoref{fig:network_arch}.
The network takes a concatenated spherical image \( I_{\textit{ref},r,x,y,z,n_x,n_y,n_z}(u,v) \) as input. We utilize backbones from the ResNet \cite{ResNet} and ShuffleNet \cite{ShuffleNet} families from torchvision \cite{torchvision2016}. The ResNet family is well-established, offering a straightforward architecture that can be efficiently adapted for embedded hardware. In contrast, the ShuffleNet family focuses on small networks with fewer parameters, providing a lightweight alternative.  As highlighted in \cite{Alonso20203DMiniNetLA}, including \( I_{x,y,z}(u,v) \) at later stages improves 3D consistency. Building on this insight, we propose that including surface normals at later stages further enhances the network's ability to identify structures by leveraging their orientation and homogeneity. Thus, we include \( I_{x,y,z}(u,v)  \)  and \( I_{n_x,n_y,n_z}(u,v) \) resized to the respective scale via nearest-neighbor interpolation to the respective stages of the backbone.
For the neck, we employ additive self-attention by using multi-scale features extracted from the backbone.
The attention-weighted features are then upsampled using deconvolution layers to the shape  \( H/2 \times W/2 \) (i. e. \( 64 \times 1024 \)). For the segmentation head, we use a deconvolution layer to reshape the output to \( H \times W \) (i. e. \( 128 \times 2048 \)), followed by two convolutional layers for anti-aliasing. For the loss functions, we use a weighted sum of the cross-entropy loss \( \mathcal{L}_{CE} \) and the Tversky loss \( \mathcal{L}_{\text{Tversky}} \) \cite{Hashemi2018TverskyAA}. The Tversky loss allows for adjusting the balance between false positives and false negatives, which is particularly useful for detecting smaller objects.

\section{\large Experiments}
\label{sec:expirement}
\begin{table*}[h]
\centering
\caption{Per class IoU (selected classes) and mIoU on scene 0006, SemanticTHAB}
\label{tab:miouresults_median}
\resizebox{\textwidth}{!}{
\begin{tabular}{|c|c|c|c|ccccccccc|c|}
\toprule
& Backbone & Parameter & Runtime & car & person & road & sidewalk & building & vegetation  & terrain & pole & traffic ind. & mIoU \\
\midrule
\multirow{1}{*}{\STAB{\rotatebox[origin=c]{0}{CENet \cite{cheng2022cenet}}}} & HarDNet & 6.8M & 63ms & 90.29 & 55.86 & 71.66 & 58.54 & 92.44 & 85.86 & 57.45 & 41.15 & 39.53 & 52.51 \\
\multirow{1}{*}{\STAB{\rotatebox[origin=c]{0}{FIDNet \cite{Zhao2021FIDNetLP}}}} & ResNet34 & 6.05M & 66ms & 91.86 & 66.18 & 69.00 & 59.24 & 91.17 & 84.06 & 41.40 & 49.64 & 57.57 & 53.58 \\
\midrule
\multirow{6}{*}{\STAB{\rotatebox[origin=c]{0}{Our}}} & ResNet18 & 14.17M & 8ms & 86.45 & 62.98 & 74.62 & 59.41 & 91.72 & 83.38 & 39.14 & 43.04 & 52.25 & 53.57 \\
    & ResNet34              & 24.27M & 12ms & 87.57 & 67.01 & 74.56 & 60.83 & 90.14 & 86.24 & 44.28 & 41.71 & 50.11 & 54.66 \\
    & ResNet50              & 65.15M & 31ms & 87.74 & 70.53 & 76.83 & 60.76 & 93.51 & 86.82 & 47.53 & 43.61 & 54.82 & 55.85 \\
    & ShuffleNet $0.5$      & 3.73M & 12ms & 88.23 & 59.44 & 73.32& 58.36 & 88.42 & 84.23 & 39.68 & 42.00 & 54.39 & 51.00\\
    & ShuffleNet $1.0$      & 9.81M & 15ms & 88.57 & 61.79 & 75.01 & 60.42 & 93.52 & 87.56 & 46.99 & 45.16& 58.59 & 53.52\\
    & ShuffleNet $1.5$      & 17.41M & 21ms & 89.29 & 67.73 & 75.83 & 59.93 & 93.35 & 87.30 & 50.34 & 46.36 & 53.77 & 55.83\\
\bottomrule
\end{tabular}
}
\end{table*}

\subsection{Experimental Setup} \label{sec:experimental_setup}
\noindent Our models was trained with a batch size of 8 and a learning rate of 0.001, utilizing the ADAM optimizer \cite{Kingma2014AdamAM}. To ensure stable training and convergence, we incorporated a learning rate scheduler. All models were initially pretrained on the SemanticKITTI dataset for 50 epochs, followed by fine-tuning on our dataset for 30 epochs. We trained our model on all sequences except 0006, which we use for testing.
We use a single Nvidia RTX 3090 for training and testing. If not explicitly stated differently, all run times are measured using a RTX 3090.

\subsection{Results} \label{sec:results}

\noindent The performance of various model architectures is summarized in \autoref{tab:miouresults_median}. We report per-class intersection over union (IoU) as well as the mean IoU (mIoU), computed for the test sequences 0006 of the SemanticTHAB dataset. Our methods demonstrate strong performance, particularly considering the complexity of the scenes in our dataset. For a fair comparison to the state-of-the-art (see \autoref{tab:benchmark_sup}), we also report the FPS and mIoU of our method (ResNet34) trained on SemanticKITTI and tested on the respective validation set using a RTX 2080 Ti at a resolution of $64 \times 512$ in \autoref{tab:benchmark_sup}. We can report comparable results by means of mIoU at a high FPS. In \autoref{tab:miouresults_median} we also report results for FIDNet and CENet which we use as baselines. We selected the baselines by means of public available and usable code (see our GitHub repo for details). Our models outperform both CENet and FIDNet by achieving higher mIoU scores (up to 55.83\%) while maintaining significantly faster run times (as low as 8ms). This makes them highly suitable for real-time applications without compromising segmentation accuracy.

\begin{figure}[!h] 
    
    \centering
    \includegraphics[width=\columnwidth]{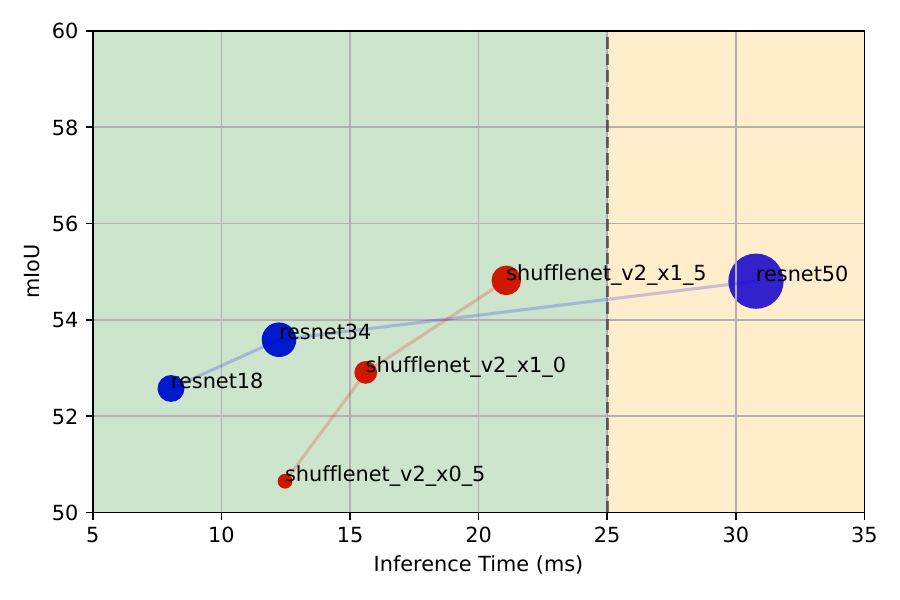}
    \caption{\textit{\textbf{Model Comparison}}: mIoU and inference time. The size of the points sketch the number of parameters.}
    \label{fig:inference_time}
\end{figure}

\noindent \autoref{fig:inference_time} shows the relationship between inference time and mIoU for different models from the ResNet and ShuffleNet families. The selection of the appropriate model depends on the specific application requirements. If inference speed is the primary consideration, ResNet18 or a smaller variant of ShuffleNet are optimal choices. Note that despite having fewer parameters, ShuffleNets exhibit slightly slower performance compared to ResNets. This discrepancy is likely due to the widespread optimization of ResNets for run time efficiency. 

\begin{figure}[!h] 
    \centering
    \includegraphics[width=\columnwidth]{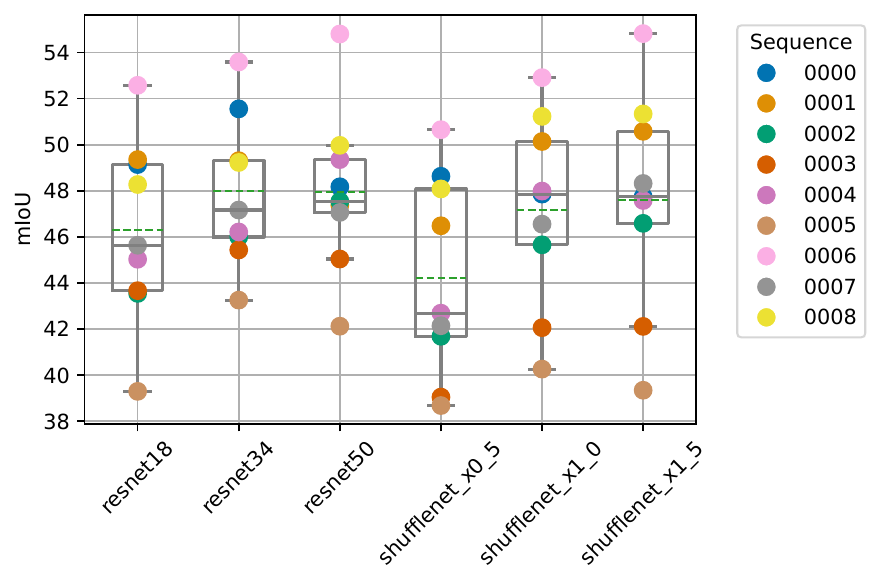}
    \caption{\textit{\textbf{Quantitative Results}}: mIoU box plot for the various sequences used in testing by cross validation.}
    \label{fig:scatter_plot}
    \vspace{-5mm} 
\end{figure}

\noindent Given the limited size of our dataset, we used leave-one-out cross-validation, training on all sequences except the one used for evaluation, to avoid overfitting and to find an unbiased network architecture that performs well across the highly diverse scenes in our dataset. The detailed results of the cross-validation for each model are presented in \autoref{fig:scatter_plot}. Based on our analysis, we recommend the use of the ResNet34 model, since it generalizes well by means of mean values (see dashed green line) across all sequences evaluated in our cross-validation, especially when considering consistency and real time constraints. 

\subsection{Ablation Study} \label{sec:ablation}
\begin{table}[h]
    \centering
    \caption{Ablation Study}
    \label{tab:SemanticLiDAR_ablation}
    \scalebox{0.85}{
    \begin{tabular}{|llll|c|c|}
    \toprule
    \(A\) &  \( M \) & \( I_{\vec{n}} \) & \( P \)  & mIoU & Increase \\
    \midrule
    \checkmark &            &            &            & 47.55 & - \\
    \checkmark & \checkmark &            &            & 47.86 & 0.65\% \\
    \checkmark & \checkmark & \checkmark &            & 49.57 & 4.24\% \\
    \checkmark & \checkmark &            & \checkmark & 53.65 & 12.83\% \\
    \checkmark & \checkmark & \checkmark & \checkmark & 54.66 & 14.95\% \\
    \bottomrule
    \end{tabular}}
\end{table}
\noindent We conducted an ablation study on the SemanticTHAB dataset to investigate the effects of pretraining \( P \), multiscale incorporation of 3D features \( M \), multiscale attention \(A\), and the use of surface normals \( I_{\vec{n}} \), as reported in \autoref{tab:SemanticLiDAR_ablation}. The evaluation is performed using the validation set, specifically sequence 0006 of the SemanticTHAB dataset. For our ablation, we train models with ResNet34 backbones for 50 epochs on the SemanticKITTI dataset to build the pre-trained models \( P \) for the respective configurations and fine-tuned them for 30 epochs on the SemanticTHAB dataset. 
\noindent The use of surface normals increases the performance to 49.57 (+4.24\%), with the best results also achieved by pre-training in SemanticKITTI, reaching 54.66 (+14.95\%). This suggests that pre-training on a related dataset (SemanticKITTI) and fine-tuning on the target dataset (SemanticTHAB) is beneficial, but its impact is maximized when combined with additional architectural components.


\begin{figure*}[!htp]
    \centering
    \begin{minipage}[b]{0.3\textwidth}
        \centering
        \includegraphics[width=\textwidth]{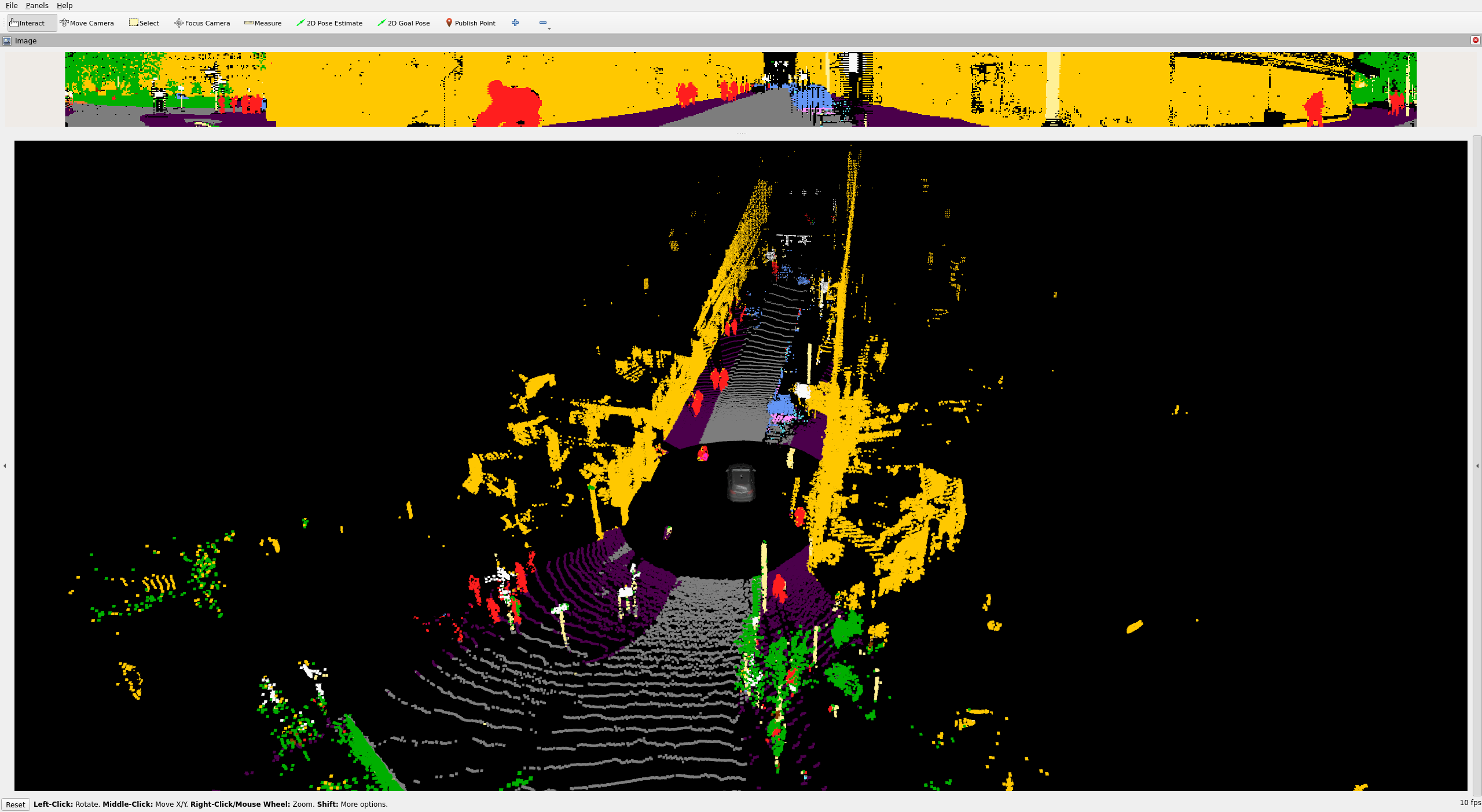}

    \end{minipage}
    \hspace{0.02\textwidth} 
    \begin{minipage}[b]{0.3\textwidth}
        \centering
        \includegraphics[width=\textwidth]{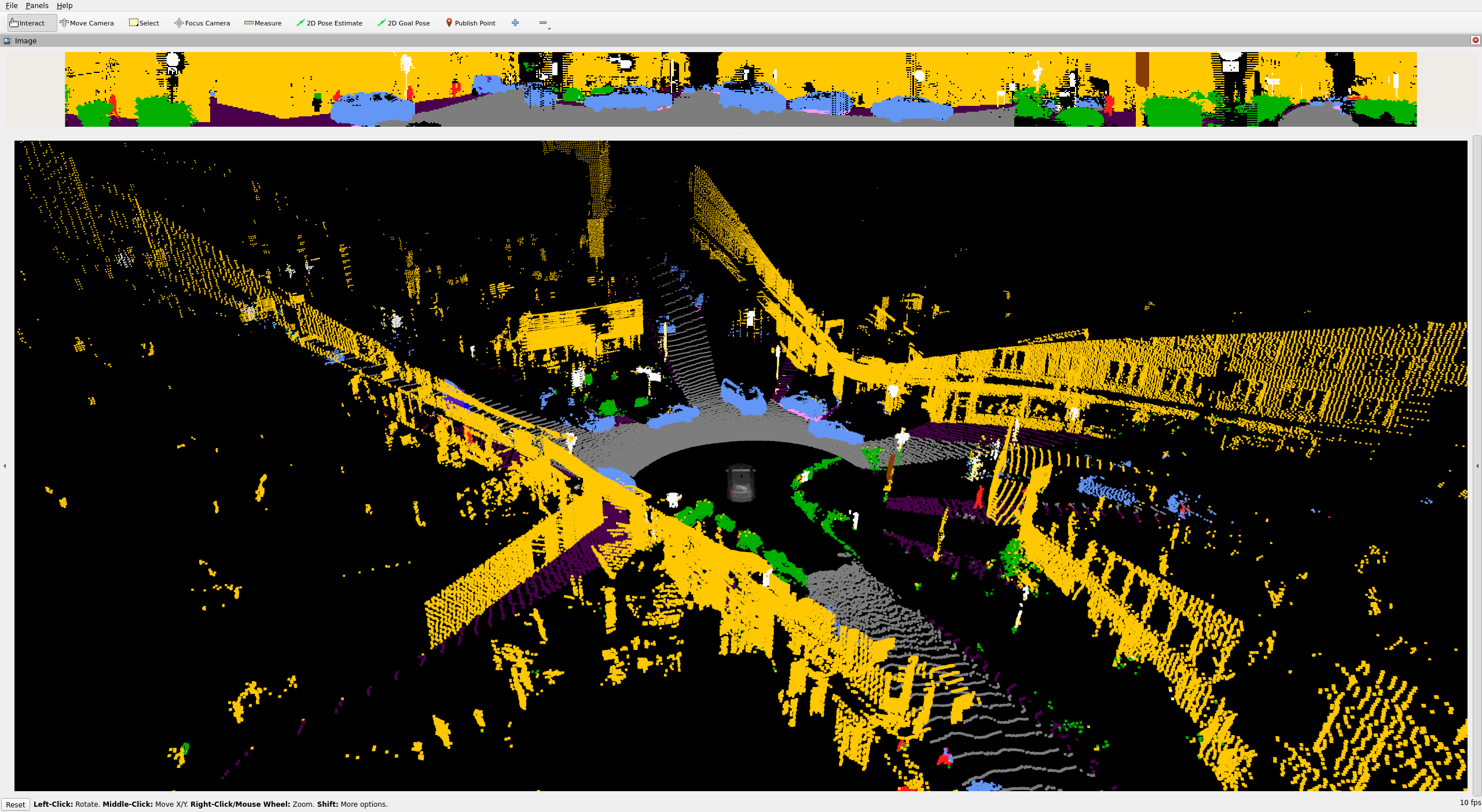}
    \end{minipage}
    \hspace{0.02\textwidth} 
    \begin{minipage}[b]{0.3\textwidth}
        \centering
        \includegraphics[width=\textwidth]{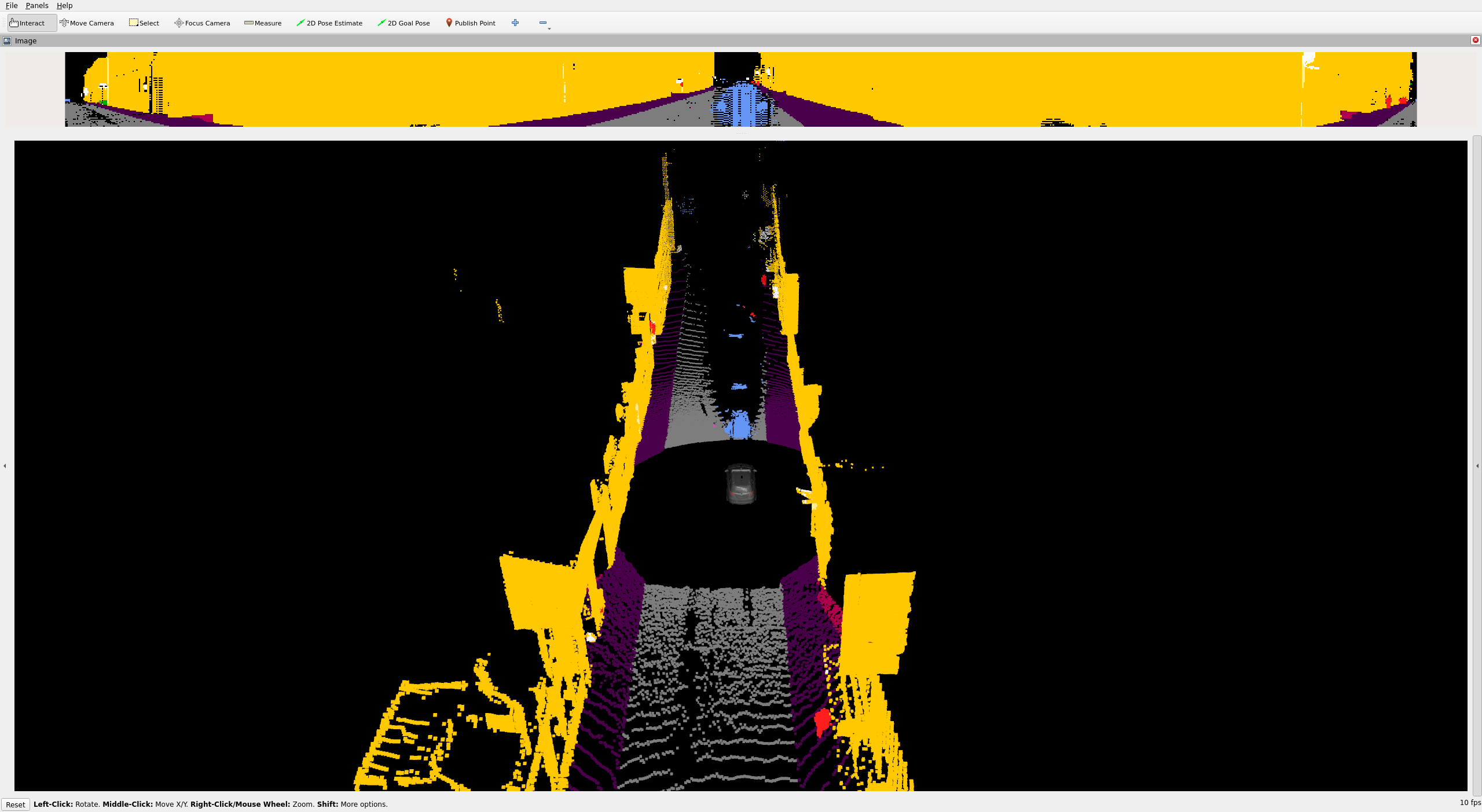}
    \end{minipage}

    \begin{minipage}[b]{0.3\textwidth}
        \centering
        \includegraphics[width=\textwidth]{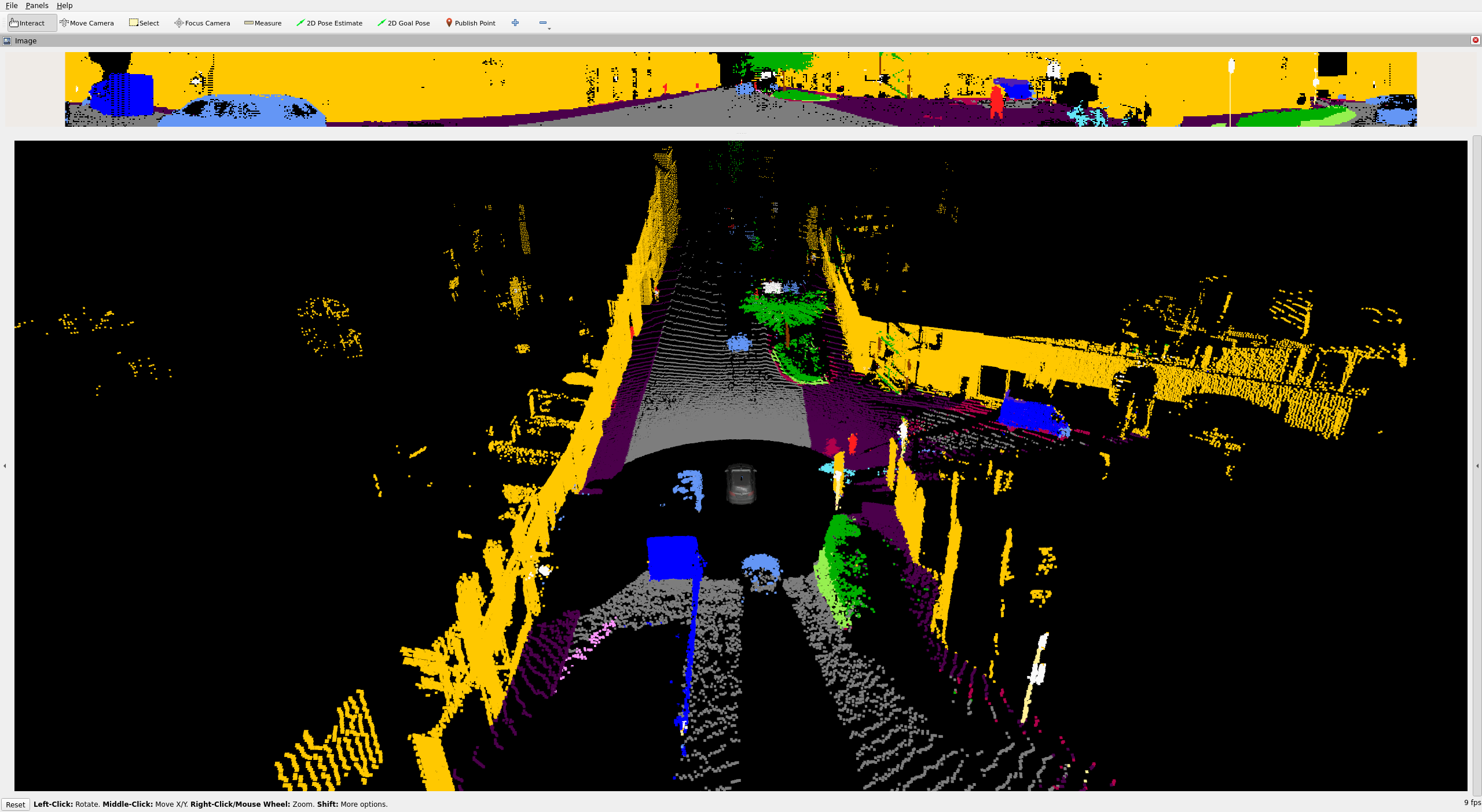}

    \end{minipage}
    \hspace{0.02\textwidth} 
    \begin{minipage}[b]{0.3\textwidth}
        \centering
        \includegraphics[width=\textwidth]{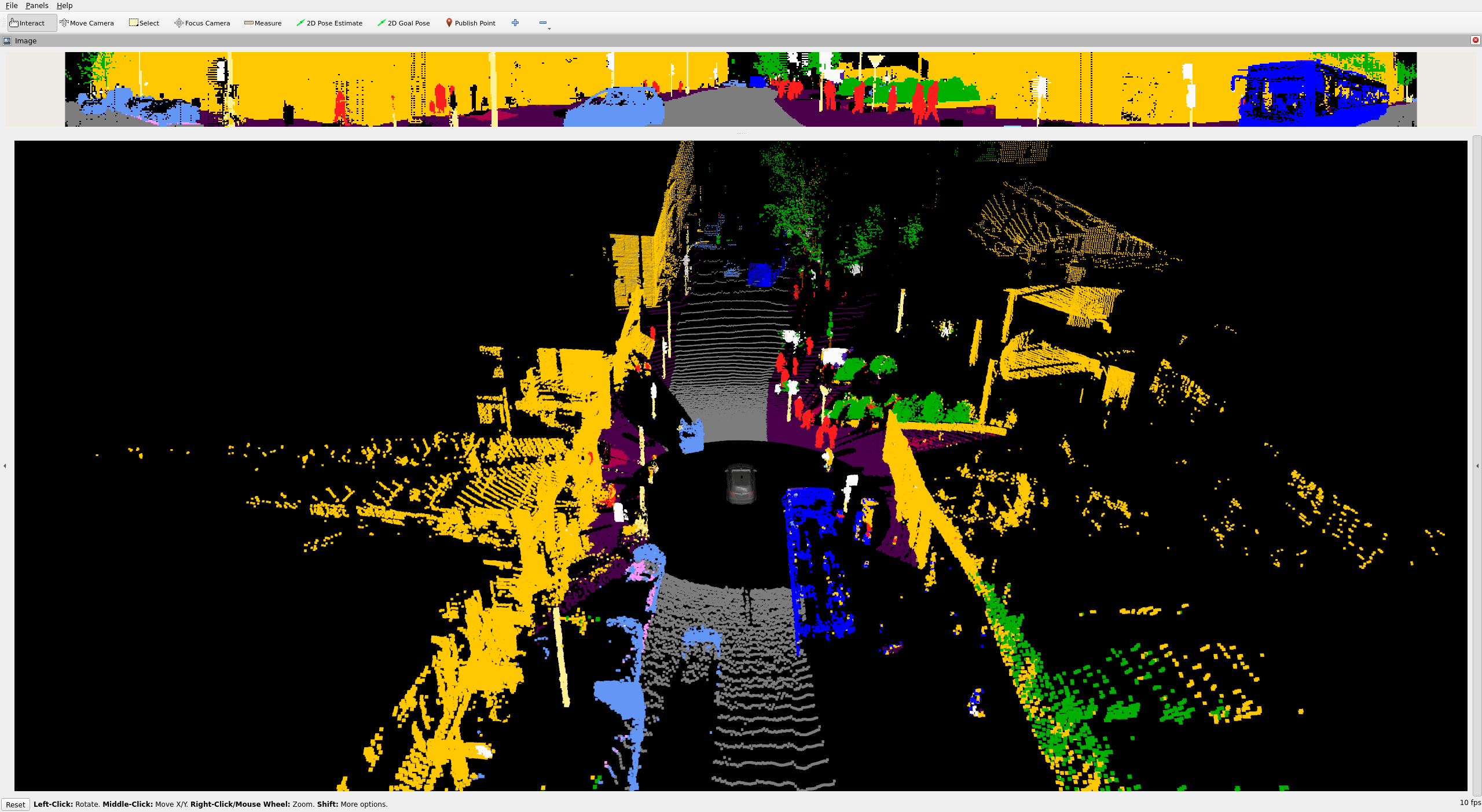}
    \end{minipage}
    \hspace{0.02\textwidth} 
    \begin{minipage}[b]{0.3\textwidth}
        \centering
        \includegraphics[width=\textwidth]{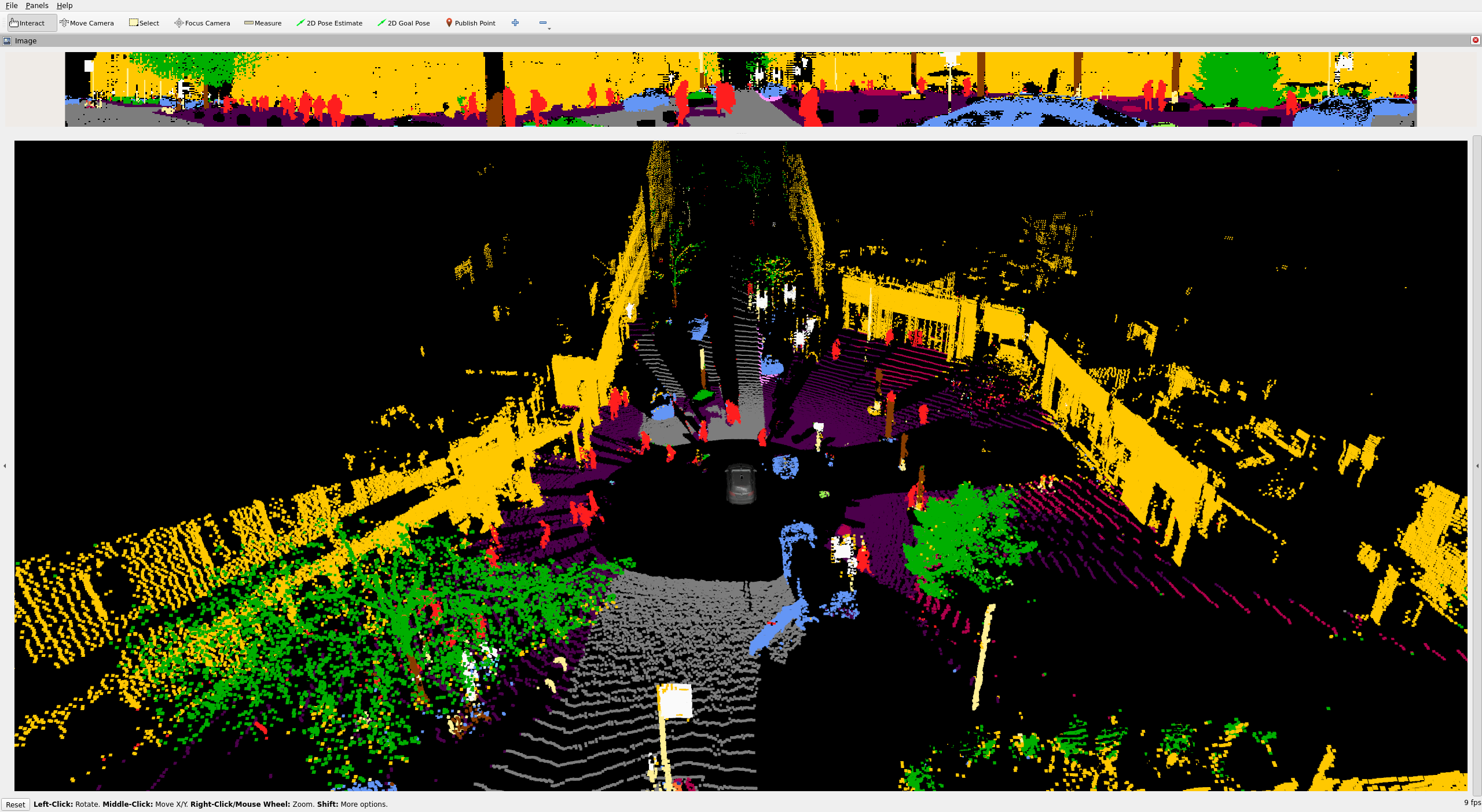}
    \end{minipage}
\caption{\textit{\textbf{Qualitative Results}}: Examples captured from our ROS2 demo. }
\label{fig:ROS2_inference}
\vspace{-5mm} 
\end{figure*}

\section{\large ROS Deployment}
\label{sec:ROS}

\noindent We implement both the training and inference code within a Docker environment. Our inference code is provided as Robot Operating System 2 (ROS) \cite{ros} nodes and deployed to a research vehicle, where it operates in real time, offering an interactive demo. The implementation is lightweight and minimizes the dependency on high-level frameworks. For deployment, we utilize a GeForce RTX 3090 GPU as it offers comparable inference performance to embedded hardware like the Nvidia DRIVE Orin given the architectural similarities and high inference throughput of both devices, making it suitable for testing with similar expected results. The qualitative results of our ROS2-based demo are shown in \autoref{fig:ROS2_inference}. For visualization, we employ RViz.

\section{\large Conclusions and Future Work}
\label{sec:conclution}
\noindent Given the availability of high-resolution LiDAR sensors today and their expected integration into products such as autonomous vehicles in the future, this work focuses on the semantic segmentation of high-resolution LiDAR scans. For this purpose, we proposed a method that projects the LiDAR point cloud into a spherical image, estimates surface normals, and utilizes CNNs for semantic segmentation of the point cloud. As part of this work, we contribute an annotated dataset derived from high-resolution LiDAR scans. To highlight the real-time capability of our method, we have developed a publicly accessible demo in ROS. Looking ahead, we expect ongoing development of LiDAR sensors with even higher resolutions. It is essential that segmentation methods evolve accordingly, and we aim to contribute to this progression with our work. In our future work, we plan to expand our data set to include a wider range of scenarios and incorporate strongly underrepresented classes, such as cyclists. Furthermore, we plan to use instance clustering methods like \cite{sautier2025alpine} to enable panoptic segmentation from our semantic segmentation approch.

{\small
\bibliographystyle{ieeetr}      
\bibliography{egbib}

\begin{thebibliography}{10}

\bibitem{behley2019iccv}
J.~Behley, M.~Garbade, A.~Milioto, J.~Quenzel, S.~Behnke, C.~Stachniss, and J.~Gall, ``{SemanticKITTI: A Dataset for Semantic Scene Understanding of LiDAR Sequences},'' in {\em Proc. of the IEEE/CVF International Conf.~on Computer Vision (ICCV)}, 2019.

\bibitem{pan2020semanticposs}
Y.~Pan, B.~Gao, J.~Mei, S.~Geng, C.~Li, and H.~Zhao, ``Semanticposs: A point cloud dataset with large quantity of dynamic instances,'' 2020.

\bibitem{jiang2020lidarnet}
P.~Jiang and S.~Saripalli, ``Lidarnet: A boundary-aware domain adaptation model for lidar point cloud semantic,'' 2020.

\bibitem{fong2021panoptic}
W.~K. Fong, R.~Mohan, J.~V. Hurtado, L.~Zhou, H.~Caesar, O.~Beijbom, and A.~Valada, ``Panoptic nuscenes: A large-scale benchmark for lidar panoptic segmentation and tracking,'' {\em arXiv preprint arXiv:2109.03805}, 2021.

\bibitem{Waymo}
P.~Sun, H.~Kretzschmar, X.~Dotiwalla, A.~Chouard, V.~Patnaik, P.~Tsui, J.~Guo, Y.~Zhou, Y.~Chai, B.~Caine, V.~Vasudevan, W.~Han, J.~Ngiam, H.~Zhao, A.~Timofeev, S.~Ettinger, M.~Krivokon, A.~Gao, A.~Joshi, Y.~Zhang, J.~Shlens, Z.~Chen, and D.~Anguelov, ``Scalability in perception for autonomous driving: Waymo open dataset,'' in {\em Proceedings of the IEEE/CVF Conference on Computer Vision and Pattern Recognition (CVPR)}, June 2020.

\bibitem{xiao20233d}
A.~Xiao, J.~Huang, W.~Xuan, R.~Ren, K.~Liu, D.~Guan, A.~E. Saddik, S.~Lu, and E.~Xing, ``3d semantic segmentation in the wild: Learning generalized models for adverse-condition point clouds,'' {\em arXiv preprint arXiv:2304.00690}, 2023.

\bibitem{FRNet}
X.~Xu, L.~Kong, H.~Shuai, and Q.~Liu, ``Frnet: Frustum-range networks for scalable lidar segmentation,'' {\em ArXiv}, vol.~abs/2312.04484, 2023.

\bibitem{Zhao2021FIDNetLP}
Y.~Zhao, L.~Bai, and X.~Huang, ``Fidnet: Lidar point cloud semantic segmentation with fully interpolation decoding,'' {\em 2021 IEEE/RSJ International Conference on Intelligent Robots and Systems (IROS)}, pp.~4453--4458, 2021.

\bibitem{cheng2022cenet}
H.-X. Cheng, X.-F. Han, and G.-Q. Xiao, ``Cenet: Toward concise and efficient lidar semantic segmentation for autonomous driving,'' in {\em 2022 IEEE International Conference on Multimedia and Expo (ICME)}, pp.~01--06, IEEE, 2022.

\bibitem{RangeFormer}
L.~Kong, Y.~Liu, R.~Chen, Y.~Ma, X.~Zhu, Y.~Li, Y.~Hou, Y.~Qiao, and Z.~Liu, ``Rethinking range view representation for lidar segmentation,'' in {\em 2023 IEEE/CVF International Conference on Computer Vision (ICCV)}, pp.~228--240, 2023.

\bibitem{RangeViT}
A.~Ando, S.~Gidaris, A.~Bursuc, G.~Puy, A.~Boulch, and R.~Marlet, ``Rangevit: Towards vision transformers for 3d semantic segmentation in autonomous driving,'' in {\em CVPR}, 2023.

\bibitem{SphereFormer}
X.~Lai, Y.~Chen, F.~Lu, J.~Liu, and J.~Jia, ``Spherical transformer for lidar-based 3d recognition,'' in {\em CVPR}, 2023.

\bibitem{Cylinder3D}
X.~Zhu, H.~Zhou, T.~Wang, F.~Hong, Y.~Ma, W.~Li, H.~Li, and D.~Lin, ``Cylindrical and asymmetrical 3d convolution networks for lidar segmentation,'' {\em arXiv preprint arXiv:2011.10033}, 2020.

\bibitem{tang2020searching}
H.~Tang, Z.~Liu, S.~Zhao, Y.~Lin, J.~Lin, H.~Wang, and S.~Han, ``Searching efficient 3d architectures with sparse point-voxel convolution,'' in {\em European Conference on Computer Vision}, 2020.

\bibitem{wu2024ptv3}
X.~Wu, L.~Jiang, P.-S. Wang, Z.~Liu, X.~Liu, Y.~Qiao, W.~Ouyang, T.~He, and H.~Zhao, ``Point transformer v3: Simpler, faster, stronger,'' in {\em CVPR}, 2024.

\bibitem{waffleiron}
G.~Puy, A.~Boulch, and R.~Marlet, ``Using a waffle iron for automotive point cloud semantic segmentation,'' in {\em IEEE/CVF International Conference on Computer Vision}, pp.~3379--3389, 2023.

\bibitem{RPVNet}
J.~Xu, R.~Zhang, J.~Dou, Y.~Zhu, J.~Sun, and S.~Pu, ``Rpvnet: A deep and efficient range-point-voxel fusion network for lidar point cloud segmentation,'' in {\em 2021 IEEE/CVF International Conference on Computer Vision (ICCV)}, pp.~16004--16013, 2021.

\bibitem{pvkd}
Y.~Hou, X.~Zhu, Y.~Ma, C.~C. Loy, and Y.~Li, ``Point-to-voxel knowledge distillation for lidar semantic segmentation,'' in {\em IEEE Conference on Computer Vision and Pattern Recognition}, pp.~8479--8488, 2022.

\bibitem{kiss-icp}
I.~Vizzo, T.~Guadagnino, B.~Mersch, L.~Wiesmann, J.~Behley, and C.~Stachniss, ``{KISS-ICP: In Defense of Point-to-Point ICP -- Simple, Accurate, and Robust Registration If Done the Right Way},'' {\em IEEE Robotics and Automation Letters (RA-L)}, vol.~8, no.~2, pp.~1029--1036, 2023.

\bibitem{2023.reichert}
H.~Reichert, M.~Hetzel, S.~Schreck, K.~Doll, and B.~Sick, ``Sensor equivariance by lidar projection images,'' in {\em 2023 IEEE Intelligent Vehicles Symposium (IV)}, pp.~1--6, 2023.

\bibitem{surface_normals}
S.~Schreck, H.~Reichert, M.~Hetzel, K.~Doll, and B.~Sick, ``Height change feature based free space detection,'' in {\em 2023 11th International Conference on Control, Mechatronics and Automation (ICCMA)}, pp.~171--176, 2023.

\bibitem{ResNet}
K.~He, X.~Zhang, S.~Ren, and J.~Sun, ``Deep residual learning for image recognition,'' {\em 2016 IEEE Conference on Computer Vision and Pattern Recognition (CVPR)}, pp.~770--778, 2015.

\bibitem{ShuffleNet}
X.~Zhang, X.~Zhou, M.~Lin, and J.~Sun, ``Shufflenet: An extremely efficient convolutional neural network for mobile devices,'' in {\em 2018 IEEE/CVF Conference on Computer Vision and Pattern Recognition}, pp.~6848--6856, 2018.

\bibitem{torchvision2016}
T.~maintainers and contributors, ``Torchvision: Pytorch's computer vision library.'' \url{https://github.com/pytorch/vision}, 2016.

\bibitem{Alonso20203DMiniNetLA}
I.~Alonso, L.~Riazuelo, L.~Montesano, and A.~C. Murillo, ``3d-mininet: Learning a 2d representation from point clouds for fast and efficient 3d lidar semantic segmentation,'' {\em IEEE Robotics and Automation Letters}, vol.~5, pp.~5432--5439, 2020.

\bibitem{Hashemi2018TverskyAA}
S.~R. Hashemi, S.~S.~M. Salehi, D.~Erdoğmuş, S.~P. Prabhu, S.~Warfield, and A.~Gholipour, ``Tversky as a loss function for highly unbalanced image segmentation using 3d fully convolutional deep networks,'' {\em ArXiv}, vol.~abs/1803.11078, 2018.

\bibitem{Kingma2014AdamAM}
D.~P. Kingma and J.~Ba, ``Adam: A method for stochastic optimization,'' {\em CoRR}, vol.~abs/1412.6980, 2014.

\bibitem{ros}
{Stanford Artificial Intelligence Laboratory et al.}, ``Robotic operating system.''

\bibitem{sautier2025alpine}
C.~Sautier, G.~Puy, A.~Boulch, R.~Marlet, and V.~Lepetit, ``Clustering is back: Reaching state-of-the-art {LiDAR} instance segmentation without training,'' {\em arxiv}, 2025.

\end{thebibliography}
}

\end{document}